\documentclass[Afour,sageh,times]{sagej}

\usepackage{moreverb,url}
\usepackage[colorlinks,bookmarksopen,bookmarksnumbered,citecolor=red,urlcolor=red]{hyperref}
\usepackage{graphicx}

\usepackage{color}
\usepackage{comment}
\usepackage{tabularx}
\usepackage{subcaption}

\usepackage{pifont}
\newcommand{\cmark}{\ding{51}}  
\newcommand{\xmark}{\ding{55}}  

\newcommand\BibTeX{{\rmfamily B\kern-.05em \textsc{i\kern-.025em b}\kern-.08em
T\kern-.1667em\lower.7ex\hbox{E}\kern-.125emX}}
\def\volumeyear{2025}

\begin{document}
\runninghead{Smith and Wittkopf}
\title{Comparing Large Language Models and Traditional Machine Translation Tools for Translating Medical Consultation Summaries -- A Pilot Study}
\author{Andy Li\affilnum{1}, Wei Zhou\affilnum{1}, Rashina Hoda\affilnum{1}, Chris Bain\affilnum{1}, and Peter Poon\affilnum{2,3}}

\affiliation{\affilnum{1} Faculty of Information Technology, Monash University, 
Clayton, VIC 3800
Australia\\
\affilnum{2} Faculty of Medicine, Nursing and Health Sciences, Monash University, 
Clayton, VIC 3800 Australia\\
\affilnum{3} Supportive and Palliative Care Unit, Monash Health, Clayton, VIC 3168 Australia\\
}

\corrauth{Rashina Hoda, Faculty of Information Technology, Monash University, 
Clayton, VIC 3800,
Australia.}

\email{rashina.hoda@monash.edu}

\begin{abstract} 
\textbf{Introduction}

This study examines the effectiveness of large language models (LLMs) and traditional machine translation (MT) tools in translating medical consultation summaries from English into the most common languages other than English spoken in Australia -- Arabic, Chinese (simplified written form), and Vietnamese. It evaluates translation quality across languages and text complexity using standard automated metrics, with a focus on healthcare applicability.

\textbf{Methods}

Two types of simulated summaries were developed: a simple summary for patients in lay language and a complex, clinician-orientated interprofessional letter including common medical jargon. Translations were produced using three LLMs (GPT-4o, LLAMA-3.1, GEMMA-2) and three MT tools (Google Translate, Microsoft Bing Translator, DeepL), and evaluated against professional third party interpreter translations  using BLEU, CHR-F, and METEOR metrics.

\textbf{Results}

Performance varied by language, model type, and summary complexity. Vietnamese and Chinese showed higher scores for the simple summary written for patients, while Arabic performed better on the complex interprofessional letter, benefiting from richer morphological context. Traditional MT tools generally outperformed LLMs on surface-level metrics, particularly for complex summaries. However, some LLMs achieved comparable or superior METEOR scores in Vietnamese and Chinese. Chinese showed the most performance decline with increased complexity.

\textbf{Conclusion}

LLMs show promise for translating into under-resourced languages like Vietnamese but remain inconsistent across contexts. Traditional MT tools offer stronger surface-level alignment, though they lack LLMs’ contextual flexibility. Current metrics fall short in capturing clinical translation quality, highlighting the need for domain-specific evaluation methods and the critical role of human oversight in ensuring responsible AI translation in healthcare. 
Future improvements could focus on fine-tuning LLMs with domain-specific medical corpora, developing safety-aware evaluation metrics, and integrating human-in-the-loop mechanisms.
\end{abstract}
\keywords{Translation, large language models, machine translation, consultation summary, responsible AI}
\maketitle

\section{Introduction}

Machine translation (MT) has seen rapid evolution in recent years, particularly with the advent of large language models (LLMs). Traditional neural MT tools, such as Google Translate \cite{googletranslate}, Microsoft Bing Translator \cite{bingtranslator}, and DeepL \cite{deepltranslator}, have been widely used in general and domain-specific applications. 
These tools rely heavily on sequence-to-sequence architectures and large-scale parallel corpora, demonstrating strong performance on high-resource language pairs, which have an abundance of digital linguistic resources available. 
However, their effectiveness is often limited in domains that require specialised vocabulary and contextual accuracy, such as medical translation~\cite{costa2012machine}.

In contrast, LLMs such as GPT \cite{openai_gpt4o}, GEMMA \cite{google_gemma2}, and LLAMA \cite{meta_llama3} have emerged as versatile alternatives capable of performing a wide range of natural language tasks \cite{María2024, Albogami2024, Lee2024, Kim2024, Ting2024, Huang2024, alam2024integrating}, including language translations \cite{wang2023document, kocmi2023large}. 
These models leverage extensive pre-training on diverse corpora, enabling them to capture broader contextual relationships and discourse-level semantics across languages. 
Previous studies have shown that LLMs can outperform traditional MT tools in document-level linguistic translation by preserving coherence and semantic intent more effectively~\cite{wang2023document}. 
Moreover, recent work has shown that LLMs can act as effective quality estimators for translation output, even without explicit references~\cite{kocmi2023large}.

Despite these advances, the application of LLMs in medical translation remains underexplored and presents unique challenges \cite{wassie2024domain, rios2024instruction}. 
Medical texts require not only accurate translation of terminology and clinical concepts but also contextual sensitivity, as errors can lead to patient harm~\cite{costa2012machine}. 
Terminological consistency, expansion of abbreviations and the handling of multilingual clinical guidelines add layers of complexity~\cite{cardey2004designing}. 
These challenges are heightened in multilingual settings like Australia, where languages such as Arabic, Chinese, and Vietnamese are widely spoken but often underrepresented in medical translation tools \cite{language}. 
The accuracy and safety of using general-purpose LLMs in such contexts remain largely untested.

This study builds on that body of research by providing an early-stage, empirical comparison of LLMs and traditional MT tools in translating palliative care consultation summaries -- a use case that is both medically sensitive and linguistically nuanced \cite{chen2024exploring}. 
Our study focuses on three of the most spoken languages other than English in Australia -- Arabic, Chinese, and Vietnamese -- each posing different linguistic and morphological challenges~\cite{language}. 
While Vietnamese, like many other languages, is relatively under-supported in traditional MT tools (e.g., not supported by DeepL), LLMs have the potential to fill these gaps through flexible, prompt-based translation. 

We designed two types of simulated summaries in English to reflect real-world use cases: a simple summary for patients,  written in lay language, and a complex, clinician-targeted version featuring domain-specific jargon, abbreviations, and nuanced medical reasoning. Translations were generated using default prompt settings for each LLM and default web version of each MT tool, and evaluated against professional third-party interpreter translations. 
Automatic evaluation metrics -- BLEU \cite{bleu}, CHR-F \cite{chrf}, and METEOR \cite{meteor} -- were used to capture surface-level overlap, morphological robustness, and semantic similarity, respectively~\cite{baek2023towards}. This multi-faceted analysis allowed us to assess the models not only in terms of fidelity to reference but also their ability to generalise across languages and document complexity.

By comparing performance across model types, summary complexity, and target languages, this study provides nuanced insights into the current capabilities and limitations of LLM-based translation for medical applications.
Our findings shed light on the importance of language-specific characteristics, the shortcomings of current evaluation metrics in capturing clinical translation accuracy, and the real-world implications of deploying LLMs in the healthcare translation workflow.
It is worth noting that real-world evaluation of LLM-based translation is essential, as translation errors in clinical communication can compromise patient safety, reduce provider efficiency, and hinder equitable access to healthcare in multilingual settings \cite{duvsek2014machine}.
This work contributes to the growing discourse on responsible and safe AI use in digital health and offers practical guidance for future research and development in this space.
\section{Methods}
This study presents an early-stage pilot comparison between popular LLMs and traditional MT tools in translating medical documents.
We selected the latest and most capable variants of three state-of-the-art LLMs: GPT-4o (OpenAI), GEMMA-2-27B (Google), and LLAMA-3.1-405B (Meta). For traditional MT tools, we selected three widely used services: Google Translate, Microsoft Bing Translator, and DeepL.
LLMs have demonstrated strong capabilities in general-purpose translation~\cite{LLM_translation_quality}, and commercial translation products based on LLMs have already been introduced into the market\footnote{https://gptforwork.com/tools/translate \\ 
  https://www.smartcat.com \\  https://languageio.com}.
However, the extent to which LLMs can be reliably and safely used for translating medical texts -- especially in safety-critical areas like digital health -- remains largely unexplored. 
This study aims to investigate this gap by evaluating translation quality across both LLMs and traditional MT tools.

\begin{table*}[h]
\centering
\caption{Translation Task Matrix (Summary Type × Language × System)}
\begin{tabular}{|l|l|c|c|c|c|c|c|}
\hline
\textbf{Summary Type} & \textbf{Language} & \textbf{GPT-4o} & \textbf{GEMMA-2} & \textbf{LLAMA-3.1} & \textbf{Google Translate} & \textbf{MS Bing} & \textbf{DeepL} \\
\hline
 & Arabic            & \cmark & \cmark & \cmark & \cmark & \cmark & \cmark \\
              Simple Summary  & Chinese           & \cmark & \cmark & \cmark & \cmark & \cmark & \cmark \\
                 & Vietnamese        & \cmark & \cmark & \cmark & \cmark & \cmark & \xmark \\
\hline
 & Arabic          & \cmark & \cmark & \cmark & \cmark & \cmark & \cmark \\
               Complex Summary  & Chinese           & \cmark & \cmark & \cmark & \cmark & \cmark & \cmark \\
                 & Vietnamese        & \cmark & \cmark & \cmark & \cmark & \cmark & \xmark \\
\hline
\end{tabular}
\smallskip

\footnotesize{\textit{Note: DeepL did not support Vietnamese translation at the time of this study. A total of 34 translations were produced across 36 combinations.}}
\label{tab:taskmatrix}
\end{table*}

\subsection{Original Summaries in English}
To simulate realistic clinical use cases, we created two types of fictitious consultation summaries in English.
The first is a short and relatively simple summary with minimal medical terminology, intended for patient-facing medical documents. 
It summarises a simulated palliative care consultation of a patient with progressive dementia who recently recovered from a urinary tract infection.
It captures the necessary components of a patient-facing summary, including medical history, recent care updates, and future recommendations.
The second summary is longer and more complex, featuring extensive medical jargon, including abbreviations and acronyms, typically used for communication purposes between clinicians (e.g., correspondence letter between the specialist and family physicians) or towards healthcare provider documents such as discharge summaries.
This simulated consultation summary also describes a much more complicated condition:
an advanced Acute Myeloid Leukemia patient who is not only unresponsive to stem cell transplantation but also has pulmonary graft-versus-host disease.
Despite multiple different treatments, pain remains opioid-refractory, requiring further evaluation and advanced interventions such as ketamine or lignocaine infusion.

It is important to emphasize that the summaries were entirely fictitious and meticulously constructed by a qualified palliative care expert with over 25 years of clinical experience. 
No personally identifiable information (PII) or sensitive patient data was included, ensuring full compliance with ethical standards and legal regulations in healthcare research.

\subsection{Selected Languages}
The target languages for translation are Arabic, Chinese (simplified), and Vietnamese, which were chosen because they are the three most commonly spoken languages other than English in Australia, as reported by the Australian Bureau of Statistics \cite{language}.
Due to DeepL not supporting Vietnamese at the time of the study, its outputs for that language were not included in our analysis.

\subsection{Reference Translations}
The reference translations for both summaries were generated by a professional third-party translation service with extensive experience in translating medical documents. 
The service is certified under ISO 17100, an international standard that establishes requirements to ensure the quality of translation services.
Two out of three translators involved were accredited by the National Accreditation Authority for Translators and Interpreters (NAATI).
Each translation underwent a rigorous quality assurance process, which included a review by an independent translator (i.e., a linguist not involved in the original translation task, providing  verification)
to identify typographical errors, formatting inconsistencies, untranslated or mistranslated content, and placeholder artifacts that may arise during document handling or post-editing.

To ensure that reference translations were reflective of industry practice and standards, we engaged accredited translation services.
However, it is common practice in the translation industry for many translators to use machine-generated drafts as a starting point.
There are no regulations that prohibit the use of machine-generated translations, and ISO 17100 also incorporates guidelines for using machine-generated translations as preliminary drafts.
Consequently, while every effort was made to ensure the quality and accuracy of the translations, some degree of influence from machine-generated drafts might still persist in the reference translations.

\subsection{Generated Translations}
For each summary, two sets of translations were generated as shown in the translation task matrix (Table \ref{tab:taskmatrix}).
The first set consisted of three LLM-generated translations.
This experiment aimed to mimic laypeople using LLM as a translation service, therefore, the basic web platform was used instead of the developer's API, which means all parameters (including system prompt) were set as the model's default values. 
All LLM-generated translations also used the same basic prompt: ``Can you translate this document into Arabic/Chinese/Vietnamese, make sure no information is lost. `\textit{DOCUMENT}' ", where \textit{DOCUMENT} is the original summary.

The second set of translations was produced using traditional MT tools -- Google Translate, Microsoft Bing Translator, and DeepL -- by inputting the same English summaries into their publicly accessible web interfaces.
As with the LLMs, no customization or advanced API usage was applied, ensuring that the outputs reflect typical usage scenarios encountered by the general public or healthcare workers relying on readily available translation tools.

\subsection{Statistical Analysis}
\begin{table*}
    \centering
    \caption{Description and limitation of each statistical metric used in the evaluation}
    \begin{tabular}{|p{0.22\linewidth}|p{0.50\linewidth}|p{0.23\linewidth}|}
         \hline
         \textbf{Metric} &  \textbf{Description} & \textbf{Limitation} \\\hline
         
         BLEU (Bilingual Evaluation Understudy) & Measures the word level of overlapping between candidate and reference translations. It's the most common metric in Machine Translation & Lacks acceptance of synonyms, paraphrasing, and overall semantic meaning \\\hline
         
         CHR-F (Character n-gram F-score) & Measures the character level of overlapping between candidate and reference translations, making it particularly effective for languages with complex morphology or segmentation issues. & Lacks acceptance of synonyms, paraphrasing, and overall semantic meaning \\\hline

         METEOR (Metric for Evaluation of Translation with Explicit ORdering) & Aligns words between candidate and reference translations, incorporating synonyms, stemming, and paraphrasing, offering a balance between surface and semantic matching, and also sensitive to syntax. & Struggles with complex semantics or meanings that require deep understanding. \\\hline
         
         
    \end{tabular}
    \label{tab:metric}
\end{table*}
We employed three widely used automatic metrics from the MT domain to obtain a comprehensive assessment for each generated summary. 
Although these metrics originate from the traditional MT domain, they remain widely accepted for evaluating LLM outputs in translation tasks \cite{wassie2024domain, elshin2024general}.
Currently, there are no standardised evaluation metrics specifically designed for LLM-generated translations, although recent studies suggest LLMs themselves may serve as quality estimators \cite{kocmi2023large}.
The metrics applied were BLEU, CHR-F and METEOR.
A brief description of each metric and their limitations are listed in Table \ref{tab:metric}. 
The values of each metric range from 0 to 1, where 1 indicates an identical matching between the reference translation and generated translation, while 0 indicates no overlapping at all.

Specifically, BLEU measures the \textit{surface similarity at the word level compared to the reference translation}. 
A high score in BLEU in the health setting means the medical terminology and linguistics for symptoms or diagnosis are captured accurately.
CHR-F measures the \textit{surface similarity at the character level compared to the reference translation}, which handles morphological variation better than BLEU.
A high CHR-F score means the generated translation is capturing medical jargon, especially abbreviations better. 
METEOR measures \textit{similarity at the semantic level}. A high METEOR score indicates that even when a medical 
condition or advice is expressed differently, the core semantic content is still preserved.
\section{Results}
\begin{figure*}
    \centering
    \begin{subfigure}[b]{0.315\textwidth}
        \includegraphics[width=\textwidth]{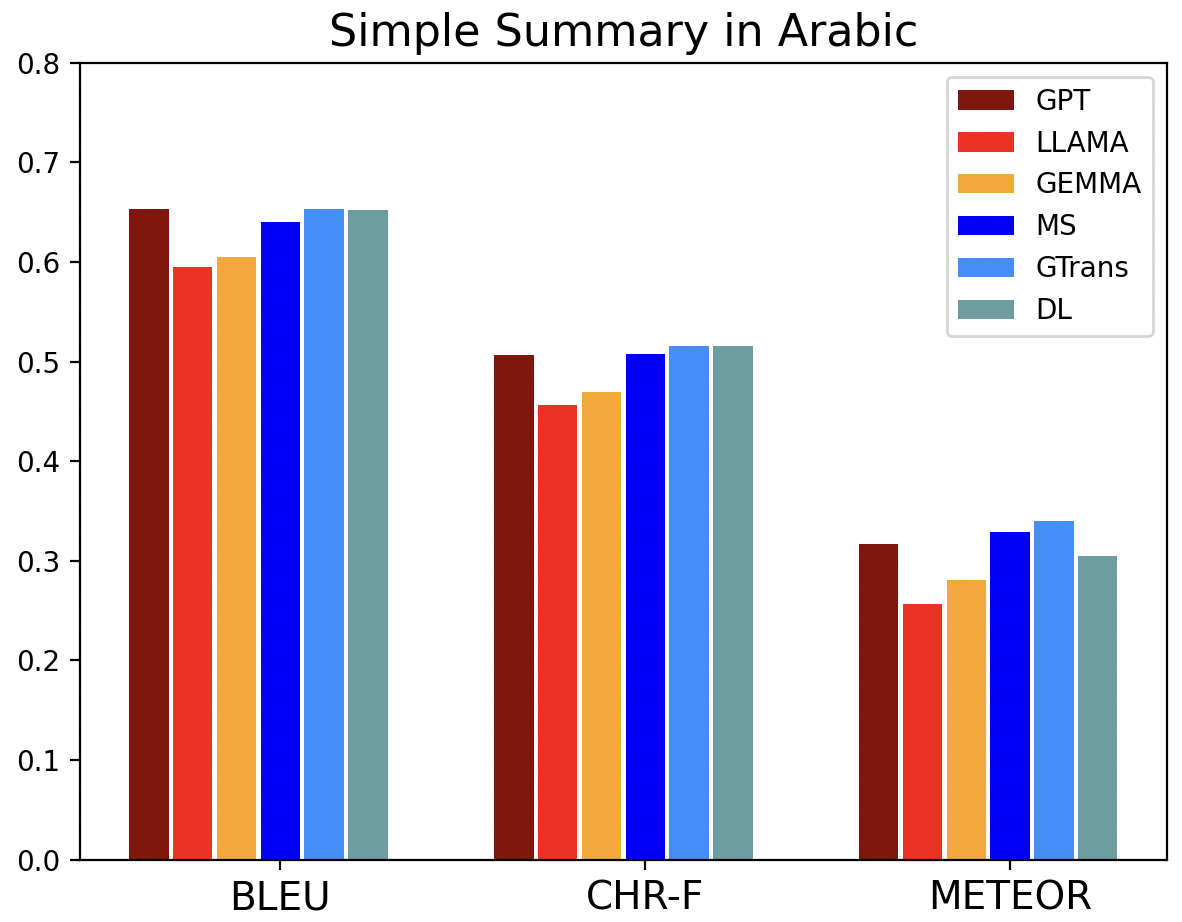}
        \caption{}
    \end{subfigure}
    \begin{subfigure}[b]{0.305\textwidth}
        \includegraphics[width=\textwidth]{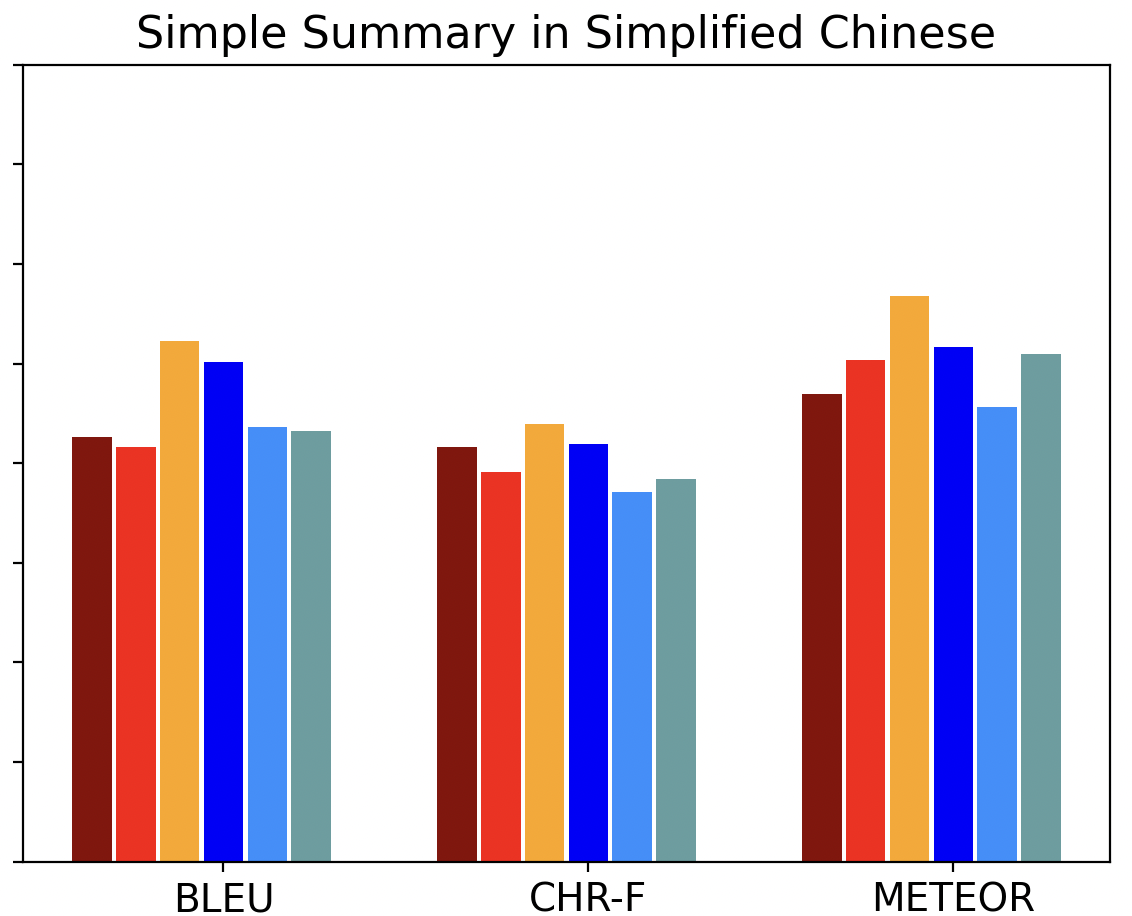}
        \caption{}
    \end{subfigure}
    \begin{subfigure}[b]{0.3\textwidth}
        \includegraphics[width=\textwidth]{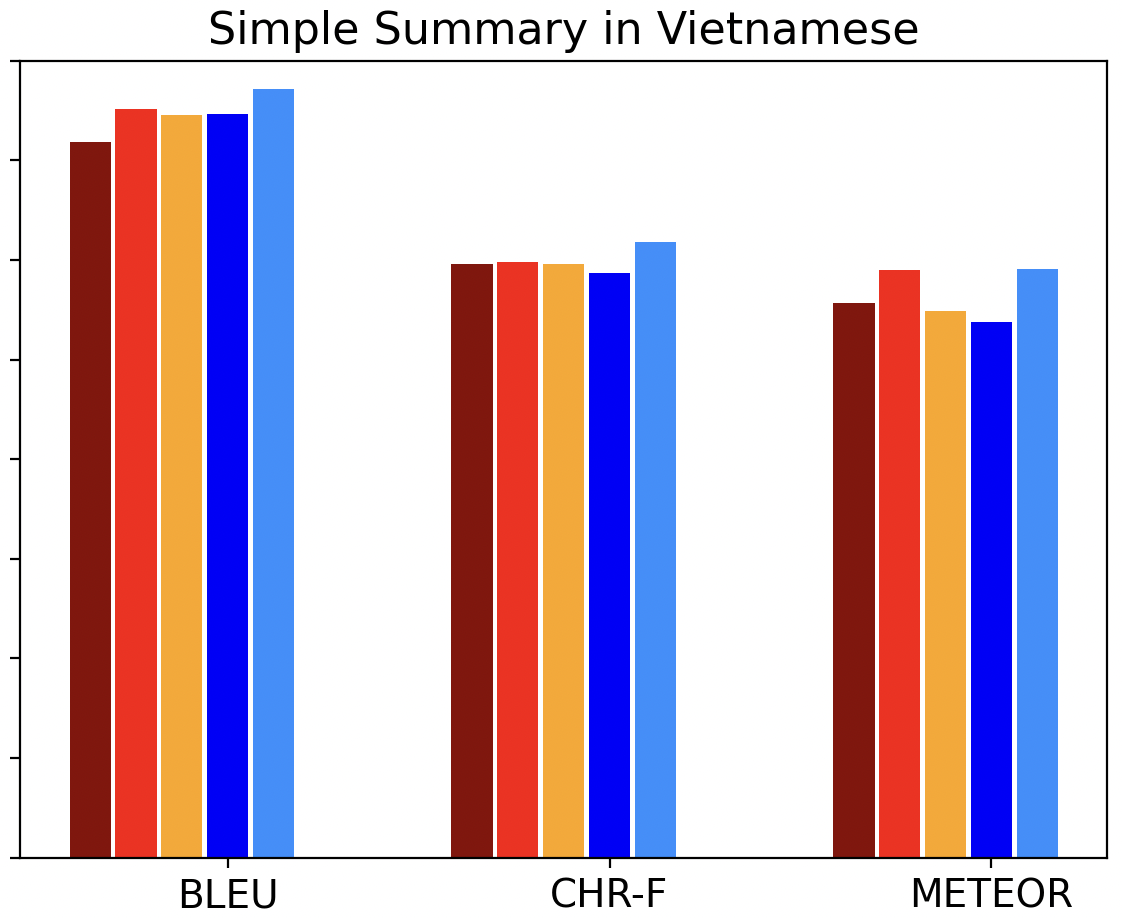}
        \caption{}
    \end{subfigure}
    \\
    \begin{subfigure}[b]{0.32\textwidth}
        \includegraphics[width=\textwidth]{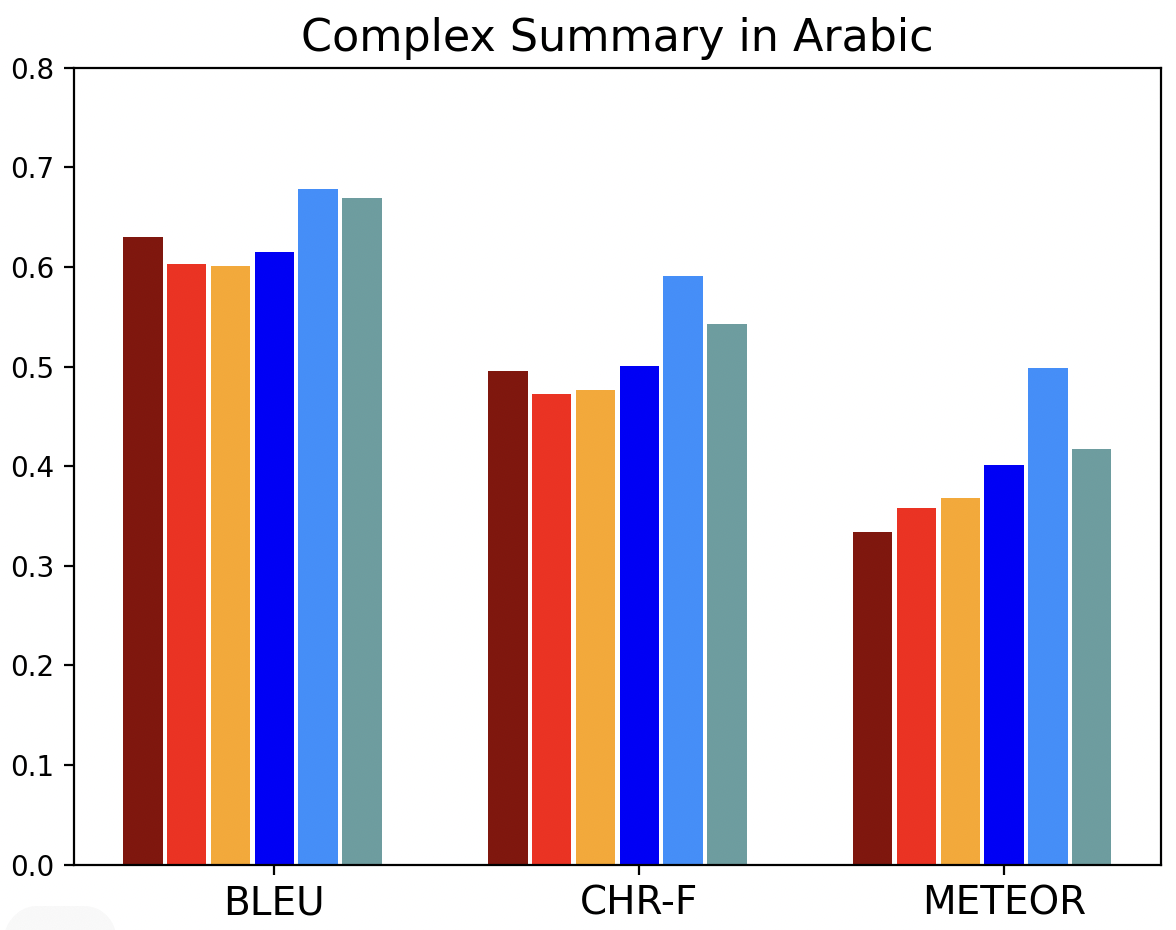}
        \caption{}
    \end{subfigure}
    \begin{subfigure}[b]{0.305\textwidth}
        \includegraphics[width=\textwidth]{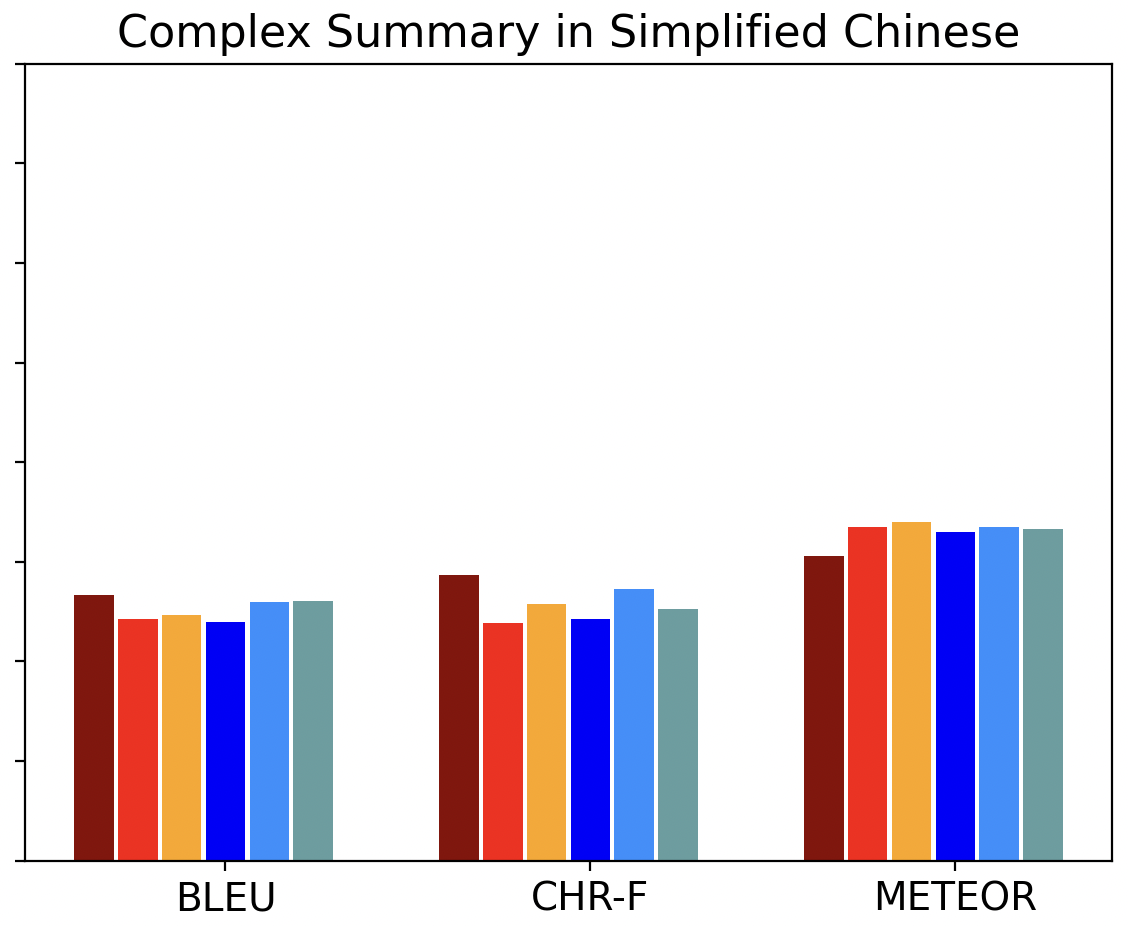}
        \caption{}
    \end{subfigure}
    \begin{subfigure}[b]{0.3\textwidth}
        \includegraphics[width=\textwidth]{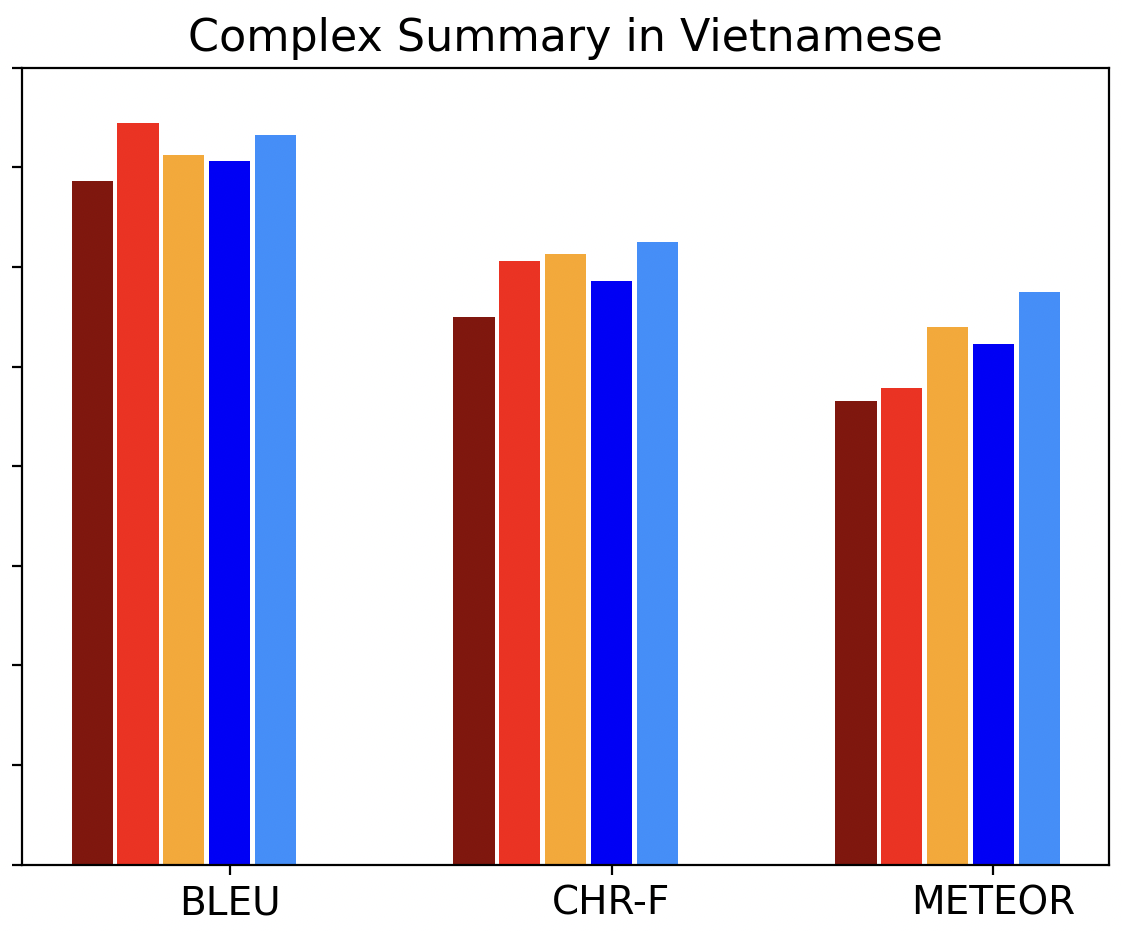}
        \caption{}
    \end{subfigure}
    \caption{Translation performance comparison across three automatic evaluation metrics -- BLEU, CHR-F, and METEOR -- for three LLMs (GPT-4o, LLAMA-3.1, GEMMA-2) and three traditional MT tools (Google Translate represented by GTrans, Microsoft Bing Translator represented by MS, DeepL represented by DL). Subfigures (a–c) show results for the simple, patient-facing consultation summary translated into Arabic, Chinese, and Vietnamese, respectively. Subfigures (d–f) show results for the more complex, jargon-rich clinical summary in the same languages. }
    \label{fig:enter-label}
\end{figure*}

Fig. \ref{fig:enter-label} presents a comparative evaluation of translation quality for two English medical consultation summaries -- simple (layperson-friendly) and complex (clinician-oriented) -- translated into Arabic, Chinese, and Vietnamese using three large language models (GPT-4o, LLAMA-3.1, and GEMMA-2) and three traditional MT tools (Google Translate, Microsoft Bing Translator, and DeepL). Translation outputs were scored using three established automatic evaluation metrics: BLEU (for surface-level n-gram overlap), CHR-F (for character-level fidelity), and METEOR (for semantic similarity and paraphrasing tolerance).

\subsection{Simple vs. Complex Summary Translation Performance}

We observed a consistent pattern across Chinese and Vietnamese where simple summaries – written in layperson language with minimal technical jargon – achieved higher scores across most metrics compared to complex summaries. This trend was less pronounced or even reversed in Arabic. For example, in Vietnamese, BLEU scores for the simple summary were high across all models, with Google Translate achieving the highest at 0.7719, followed closely by LLAMA (0.7517) and GEMMA (0.7458). CHR-F and METEOR also reflected strong alignment, with Google Translate again leading in METEOR (0.5908). 

However, for the complex summary, BLEU scores dipped across all models (e.g., GPT declined from 0.7188 to 0.6857), though the degradation for Vietnamese was relatively minor, suggesting good resilience for Vietnamese translation. 
A dramatic drop was observed in Chinese performance. For instance, GEMMA's BLEU fell from 0.5227 (simple) to 0.2466 (complex). CHR-F and METEOR scores showed a similar drop, with CHR-F dropping from 0.4392 to 0.2576, and METEOR from 0.5678 to 0.3394. This suggests a struggle in translating complex, technical content due to syntactic and terminological challenges.

In Arabic, an interesting reverse trend emerged. For instance, Google Translate's BLEU score increased from 0.6528 (simple) to 0.6787 (complex). METEOR also jumped significantly, from 0.3399 to 0.4988. This may be due to Arabic’s rich morphology, where longer, more context-rich sentences provide better clues for disambiguation and grammatical accuracy.

\subsection{LLMs vs. Traditional MT Tools}
Across all languages, traditional MT tools (Google Translate, Microsoft Bing Translator) generally outperformed LLMs like GPT-4o, LLAMA, and GEMMA on standard metrics, particularly for complex summaries.

For example, in Arabic complex summaries, Google Translate achieved a BLEU of 0.6787 and METEOR of 0.4988, outperforming GPT (BLEU: 0.6300, METEOR: 0.3343), LLAMA, and GEMMA. Similar trends were seen in CHR-F, where Google Translate scored 0.5907, the highest among all systems.

This reflects the focus of traditional MT tools on token-level alignment and training objectives that favour metrics like BLEU and CHR-F, rewarding close n-gram matches. On the other hand, LLMs often prioritise fluency and coherence, producing paraphrased outputs that, while semantically accurate, may score lower due to structural differences.

However, LLAMA outperformed GPT-4o in Vietnamese and Chinese METEOR scores, and GEMMA did the same except for Vietnamese simple summaries.
For example, in Vietnamese simple summaries, LLAMA achieved 0.5894 METEOR, slightly higher than GPT (0.5566). This suggests some LLMs may capture semantic similarity better in certain contexts.

\subsection{Language-Specific Performance Trends}
The three target languages demonstrated distinct translation characteristics in response to summary complexity and model type:
\begin{itemize}
    \item Arabic: Translation quality improved with complex summaries across most models. For instance, GPT's CHR-F went from 0.5064 (simple) to 0.4957 (complex) – a small drop, but METEOR rose from 0.3166 to 0.3343, and other models (such as LLAMA and GEMMA) saw larger METEOR gains. This supports the hypothesis that longer, more redundant input helps Arabic models resolve morphological ambiguities.
    \item Chinese: Performance dropped sharply from simple to complex summaries. BLEU scores for GEMMA fell from 0.5227 to 0.2466, and METEOR from 0.5678 to 0.3394. This suggests transliteration issues, syntactic mismatches, and rare term handling present challenges in Chinese clinical text translation. 
    Similar challenges in Chinese medical translation have been reported in prior studies, where syntactic ambiguity, segmentation issues, and limited domain-specific parallel corpora reduce translation accuracy \cite{cardey2004designing}.
    \item Vietnamese: Scores declined modestly from simple to complex, showing the least performance degradation. For example, LLAMA’s BLEU only dropped from 0.7517 to 0.7441, and CHR-F actually increased from 0.5979 to 0.6055. This suggests that Vietnamese translation is robust, possibly due to better multilingual representation in modern models.
\end{itemize}
\section{Discussion}
\subsection{Limitation of Evaluation Metrics}
Although this study reports a variety of automatic evaluation metric scores, the results should only serve as an indicative assessment and relative comparison, rather than a definitive evaluation of translation quality.
Translation quality evaluation remains inherently challenging due to the absence of a single ground truth, as different translations can be equally valid while varying in structure, wording, and style. 
A major limitation of existing metrics including those we used in this paper is their inability to capture the deeper linguistic and contextual aspects of translation, particularly in cases involving idioms, sarcasm, and cultural nuances. 
These elements often require interpretation rather than direct word-for-word translation, meaning that a sentence that appears different from a reference translation could still be equally or even more accurate in conveying the intended message.
Moreover, context plays a crucial role in determining the appropriateness of a translation, as words and phrases can shift in meaning depending on discourse, tone, or cultural expectations. 
In the case of specialised domains like medical documents, metrics also fail to account for the critical importance of certain terms, such as medicines or medical conditions, where a mistranslation can have serious consequences \cite{vieira2021understanding, eshbo2024problems, ismayilli2024navigating}.
Conversely, missing a less significant word, such as a filler word, typically has little impact on overall meaning, however, they receive the same penalty.  
As a result, existing metrics provide only an approximation of translation quality.
They often fail to reflect accuracy on critical content, fluency, coherence, and true semantic equivalence, making evaluation scores an imperfect measure of overall translation accuracy. 
Moreover, MT-based metrics such as BLEU and CHR-F may penalise translations that are semantically accurate but structurally different from the reference, especially in paraphrased or culturally adapted expressions. In medical contexts, these limitations become critical, as mistranslating a drug name carries more risk than omitting a filler word -- yet both receive equal penalty under most metrics \cite{costa2012machine}.

\subsection{Limitation of LLMs}
LLMs for medical translation face several key limitations. 
First, they may inaccurately map medical terminology, leading to critical mistranslations, such as confusing medication names or dosage instructions. 
Second, LLMs sometimes generate ambiguous translations, especially when interpreting symptoms or clinical terms, which can affect the accuracy of medical advice. 
Additionally, while Chinese has extensive linguistic data available for training, Arabic \cite{lowResourceArabic} and Vietnamese \cite{lowResourceVi,lowResourceVi2} remain relatively low-resource languages, especially in medical contexts.
This means that LLMs trained primarily in English and other high-resource languages may produce lower-quality translations for Arabic and Vietnamese due to insufficient medical data, leading to inconsistent terminology and missing critical nuances. Lastly, translation quality heavily depends on the input prompt \cite{zhang2023prompting, garcia2023unreasonable}, so minor differences in phrasing or context can result in significant variations in output, making it difficult to evaluate the overall capability of an LLM model for medical translation. 

Given the growing popularity of LLMs, many clinicians, particularly those in non-government private practices, are already using them in clinical practice for various purposes, such as reducing administrative burdens, enhancing diagnostic accuracy, and personalising treatments \cite{blease2024generative, american2023physician}.
Patients are also seen to use LLMs without any professional medical knowledge to translate medical documents that may contain critical information, including medications and medical advice \cite{aydin2024large}.

This research highlights a significant gap between current academic studies on the responsible and safe use of LLMs in healthcare and how these tools are actually being used in practice. 
It is crucial to raise awareness from both an academic and practical perspective: more research is needed to better understand the implications for patient safety, and guidelines should be developed to support existing clinical practices. 
At the same time, clinicians and patients must be made aware of the potential risks of using LLMs without the involvement of qualified healthcare professionals.

These challenges reinforce the need for a `human-in-the-loop' approach, where clinical experts review and validate translations before use in patient care. Such oversight mitigates risks and aligns LLM outputs with professional standards of safety and accountability.

\section{Conclusion}
The findings of this comparative pilot study underscore both the strengths and limitations of LLMs and traditional MT tools in the context of medical document (patient consultation summary) translation for the three most widely spoken foreign languages in Australia.
In addition, this paper discusses the challenges inherent in quantitatively evaluating translation quality, particularly with regard to the reliance on automatic metrics.
It also addresses specific limitations associated with employing LLMs for the translation of medical documentation.

Despite these challenges, LLMs are already being integrated into real-world translation workflows. Professional translators increasingly use LLM-generated drafts as a starting point, refining them for accuracy, while lay users often rely on these translations without verification. This dual usage underscores the reality that LLM-based medical translation is no longer a question of if it will be used, but rather how it should be used responsibly.

Given the inevitable use of LLMs in translation and its potential risks and benefits, the key issue moving forward will be how they can be best leveraged to complement human expertise while minimizing errors. As LLMs continue to evolve, future work can focus on resourcing under-resourced languages better and developing frameworks that ensure their medical translations are both accurate and ethically sound, particularly for users who lack the expertise to critically evaluate their reliability.

\section{Data Availability Statement}
The authors declare that all the data supporting the findings of this study are available
online via the \href{https://github.com/AndyLi-26/Medical-Summary-Translation}{link}.

\section{Declaration of conflicting interests}
The authors declared no potential conflicts of interest with respect to the research, authorship, and/or publication of this article.

\section{Funding}
The cost of translations by certified human translators and article processing was supported by the Faculty of Information Technology, Monash University.


\section{Ethical approval and informed consent statements}
Ethical approval and patient consent were not applied in this study because no actual patient or doctor data was used.

\section{Acknowledgements}
No third-party submissions or writing assistance were used.

\section{Guarantor}
Andy Li

\section{Contributorship}
Andy designed the experiments with inputs from Wei, Rashina, and Peter. Peter wrote the original English summaries. Andy performed the experiments and analysed the data. 
Wei assisted with the measurements. 
Andy and Wei wrote the manuscript in consultation with Rashina, Peter and Chris.
All authors discussed the results and contributed to the final manuscript.

\bibliographystyle{plainnat}
\bibliography{bib}

\end{document}